\documentclass[a4paper,10pt,twocolumn]{article}
\usepackage[pdftex]{graphicx}
\usepackage[top=30truemm,bottom=30truemm,left=20truemm,right=20truemm]{geometry}
\usepackage{amsmath,amssymb}

\title{\bfseries Non-expert to Expert Motion Translation Using Generative Adversarial Networks\thanks{This paper is based on a presentation at the International Symposium on Applied Abstraction and Integrated Design (AAID2023), held on March 4--5, 2023.}}
\author{
Yuki Tanaka \\
Keio University \\
Yokohama, Japan \\
tanaka@katsura.sd.keio.ac.jp
\and
Seiichiro Katsura\\
Keio University \\
Yokohama, Japan \\
katsura@sd.keio.ac.jp
}

\date{}

\begin{document}

\maketitle
\thispagestyle{empty}

\begin{abstract}
Decreasing skilled workers is a very serious problem in the world. To deal with this problem, the skill transfer from experts to robots has been researched. These methods which teach robots by human motion are called imitation learning. Experts' skills generally appear in not only position data, but also force data. Thus, position and force data need to be saved and reproduced. To realize this, a lot of research has been conducted in the framework of a motion-copying system. Recent research uses machine learning methods to generate motion commands. However, most of them could not change tasks by following human intention. Some of them can change tasks by conditional training, but the labels are limited. Thus, we propose the flexible motion translation method by using Generative Adversarial Networks. The proposed method enables users to teach robots tasks by inputting data, and skills by a trained model. We evaluated the proposed system with a 3-DOF calligraphy robot. 
\end{abstract}

\section*{Keywords}
Skill transfer, motion-copying system, machine learning, generative adversarial networks

\section{Introduction}

In recent years, decreasing working population is a very serious problem in the world. Especially, skilled workers are declining against a lot of needs for delicate tasks. To deal with this problem, the skill transfer from experts to robots is one of the solutions. Teaching robots tasks by human motion is generally called imitation learning. Imitation learning has been researched with various methods~\cite {fang,hussein}. Some research uses machine learning to model the relationship between environmental image and motion data~\cite{johns, yang}. Yang et al. succeeded in finishing folding various fabric objects against interference by a third person. Others calculated the warp function which adapts demonstrated trajectory to the current scene~\cite{lee,huang}. They used Dynamic Time Warping (DTW)~\cite{sakoe} as the warping function. They succeeded in tying a knot, folding a towel, erasing a
whiteboard, and tying a rope when the property of objects is changed. This research aims to reproduce the demonstrated motion according to environmental changes. In these methods, the trained model corresponds one-to-one with the task description. Thus, when operators want to teach robots various tasks, models need to be trained conditionally. 

As can be seen in the above research, humans, especially experts' skills generally appear in not only position data, but also force data. Thus, position and force data need to be saved and reproduced. A lot of research has been conducted in the framework of motion-copying systems~\cite {yokokura}. The motion-copying system consists of two robots, a leader and a follower. Bilateral control~\cite{matsumoto} is implemented between them to synchronize position and realize the law of action and reaction. Under bilateral control, operators can teach robots tasks while touching the environment remotely. The operator's position and force data are saved and reproduced as a virtual leader robot. Motion-copying systems can reproduce saved motion adaptively and accurately~\cite{nishimura,kobayashi} by cooperating with humans and estimating task-dependent parameters. However, the task cannot be changed in the reproduction phase. To solve this problem, recent research uses machine learning to generate motion~\cite{hayashi,sakaino,saigusa}. They used Long Short-Term Memory (LSTM) unit to approximate the relationship between the position and force of two robots. In their method, the control period does not get worse when LSTM is implemented in the control system. However, task specification is discretized and limited.   

Thus, we propose the flexible motion translation method by using Generative Adversarial Networks (GAN). The proposed method enables users to teach robots tasks by input motion, and skills from a trained model. We evaluated the proposed system with a 3-DOF calligraphy robot. 
\section{Preliminary Knowledge}
\subsection{Generative Adversarial Networks}

\begin{figure}[t]
    \begin{center}
      \includegraphics[width=8cm]{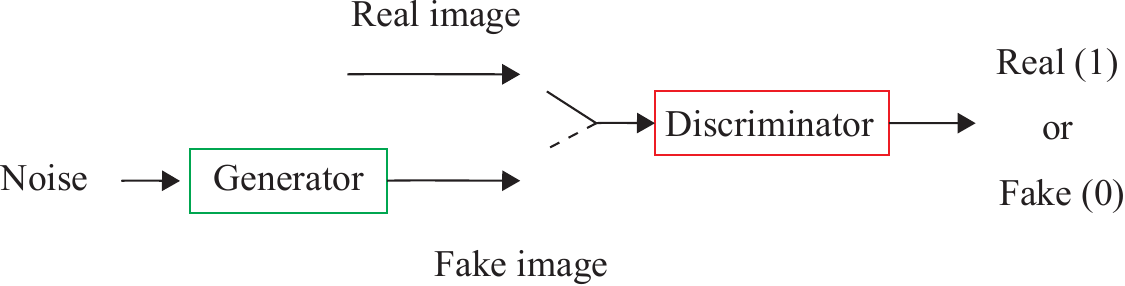}
      \caption{Structure of GAN for image generation.}
      \label{fig:gan}
    \end{center}
\end{figure}

In this section, we explain the basic structure of GAN~\cite{goodfellow} and the image translation method, Pix2Pix~\cite{isola}. GAN is the one of generative models in machine learning fields. GAN has been used in various academic fields such as voice and text generation~\cite{yang02,nie}. Especially, GAN has succeeded in generating images, and some developed GAN-based methods have been proposed~\cite{radford,karras}. GAN consists of two networks, generator $G$ and discriminator $D$. A structure of GAN is shown in Fig. \ref{fig:gan}. $G$ aims to deceive $D$ and generates the fake data. $D$ aims to determine whether input data is true or fake data. When input data is true, $D$ should output $1$. Otherwise, when input data is fake, $D$ should output $0$. Their goals conflict with each other, thus, they are trained competitively. Its training process is expressed by
\begin{equation}\label{gan_loss}
\begin{split}  
    \min _G \max _D V(D, G) &=\mathbb{E}_{\boldsymbol{x} \sim p_{\text {data }}(\boldsymbol{x})}[\log D(\boldsymbol{x})]\\
    &+\mathbb{E}_{\boldsymbol{z} \sim p_{\boldsymbol{z}}(\boldsymbol{z})}[\log (1-D(G(\boldsymbol{z})))]        
\end{split}
\end{equation}
where $x$ and $z$ represent true data and noise vector respectively. For $D$, the first term on the right hand of the Eq. (\ref{gan_loss}) must be maximized because the input is true data. The second term on the right hand of it must be maximized because the input is fake data. For $G$, the first term on the right hand of the Eq. (\ref{gan_loss}) is neglected because outputs of $G$ are unrelated. The second term on the right hand of it must be minimized because fake data should disguise as true data. $z$ is playing the role of seed value of $G$, so outputs are different from each other.  

For simplicity, the loss calculated by Eq. (\ref{gan_loss}) is rewritten as binary cross entropy loss $L_{\mathrm{BCE}}$ by
\begin{align}\label{bce}
    L_{\mathrm{BCE}}=\mathbb{E}[t\log D(\boldsymbol{x})+(1-t)\log (1-D(\boldsymbol{x}))]
\end{align}
where $t$ and $x$ represent the answer label and input data of DNN. When $t$ is $1$, $L_{\mathrm{BCE}}$ equals the first term on the right hand of the Eq. (\ref{gan_loss}). When $t$ is $0$, $L_{\mathrm{BCE}}$ equals the second term on the right hand of the Eq. (\ref{gan_loss}). 

\subsection{Image Translation}

As one of the developed GAN, image translation methods, Pix2Pix has been proposed. Pix2Pix can translate images such as a horse to zebra, and line drawing to real pictures. In the Pix2Pix model, $z$ is replaced by a image. Against general GAN, inputs of $D$ are not only output of $G$ and true data, but also pre-translated images. This technique suppresses generating images not considering pre-translated image. Through training of gan, $D$ is a dynamic loss function. However, training two related networks is difficult. In Pix2Pix, $L_1$ loss between generated and true data is considered. $L_1$ loss is calculated by
\begin{align}\label{L1}
    L_1=\mathbb{E}[||y- G(\boldsymbol{x})||_1].
\end{align}

$L1$ loss is too simple to solve complex tasks. Thus, we need to adjust the scale of $L_1$. The overall loss of $G$ is calculated by
\begin{align}\label{loss_all}
    L=L_{BCE}+\lambda L_1
\end{align}
where $\lambda$ is the hyperparameter of training. We need to implement some dropout layer in the network of $G$ because Pix2Pix uses no noise vector $z$. Dropout is a method that stops some nodes' outputs in the training phase to realize a more generative model. By introducing these equations and some techniques, Pix2Pix succeeded in translating images. 

\section{Proposed Method}

\begin{figure}[t]
    \begin{center}
      \includegraphics[width=8cm]{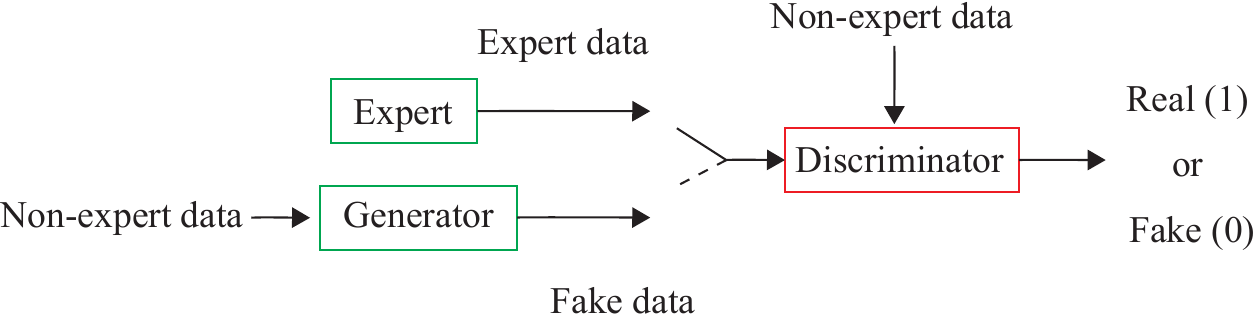}
      \caption{Structure of the proposed method.}
      \label{fig:mo2mo}
    \end{center}
\end{figure}

\begin{figure}[t]
    \begin{center}
      \includegraphics[width=7cm]{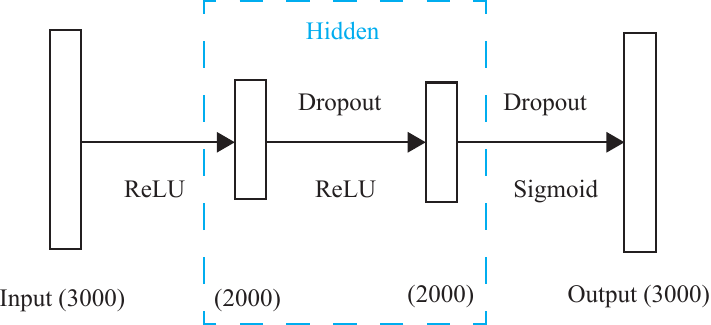}
      \caption{Structure of the generator.}
      \label{fig:gen}
    \end{center}
\end{figure}

\begin{figure}[t]
    \begin{center}
      \includegraphics[width=7cm]{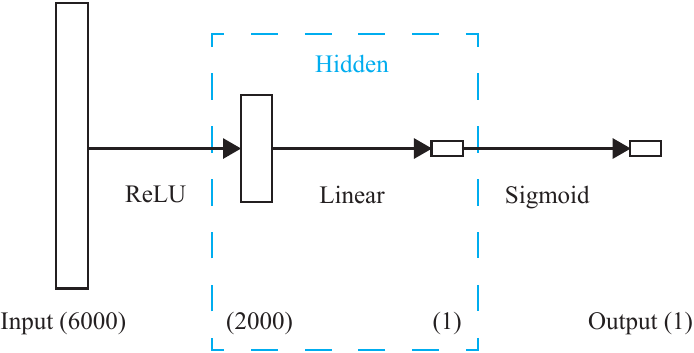}
      \caption{Structure of the discriminator.}
      \label{fig:dis}
    \end{center}
\end{figure}

\begin{figure}[t]
    \begin{center}
      \includegraphics[width=8cm]{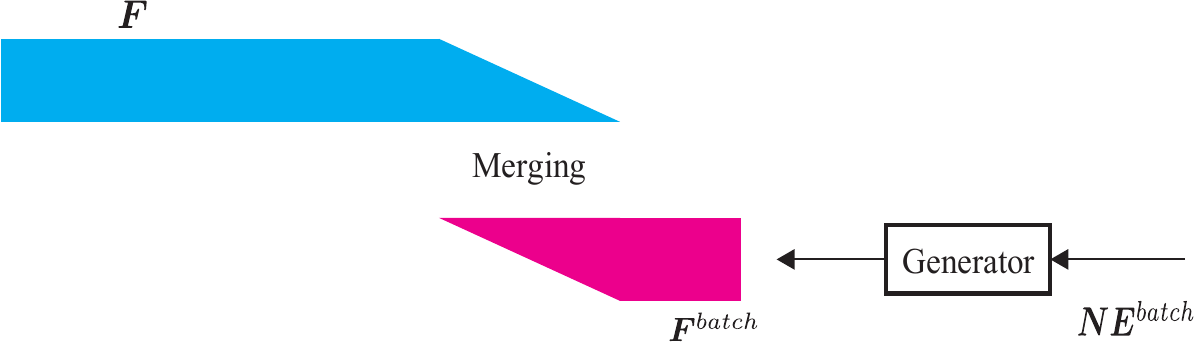}
      \caption{Time series merging method.}
      \label{fig:merge}
    \end{center}
\end{figure}

\subsection{Preprocessing and translation}
In this section, we explain the motion translation system. The structure of the proposed method is shown in Fig. \ref{fig:mo2mo}. Pre-translated images and real images in Pix2Pix are replaced by non-expert and expert motion data, respectively. The proposed system aims to enable users to teach robots tasks by non-expert data, and skills by expert data. Similar to Pix2Pix, paired motion is needed in the proposed method. However, it is not enough to match non-expert and expert motion at the same time because the pace and completion time of motion are different. To solve this problem, we used DTW. DTW is one of dynamic programming which can calculate the minimum path between two time series. Manhattan and euclidean distance are also used to check the similarity between two time series, however, they cannot be used for data that has a different period or data size. DTW is effective to these data.

Examples of DNN models in $G$ and $D$ are shown in Figs. \ref{fig:gen} and \ref{fig:dis}. In this model, the input and output of $G$ are 3-DOF position and force data, and the input of $D$ is a pair of 3-DOF motion data. The output layer includes a sigmoid function because input data are normalized. Output motion data is modified to the size of pre-translated data by using max and min values. For other activation functions, we chose Rectified Linear Unit (ReLU).

\subsection{Post-processing}
In addition, DNN only generates fixed-size motion data, thus, we need to reconstruct the full-sized motion data after that. We used a merging method shown in Fig. \ref{fig:merge}. The full-sized motion data is reconstructed by
\begin{align}\label{merge}
  F_i = \frac{|F|-i}{nd}F_i + (1-\frac{|F|-i}{nd})F^{\mathrm{batch}}_{i-|F|+nd}
\end{align}
where $F$, $n$, and $d$ denote generated motion data, sample size, and window step size, respectively. When $i$ equals $|F|$, $F^{\mathrm{batch}}$ only remains. On the other hand, when $i$ equals $nd-|F|$, $F_i$ only remains. This method can merge two time series data without peak occurrence. 
\section{Experiments}

\begin{figure}[t]
  \begin{center}
    \includegraphics[width=8cm]{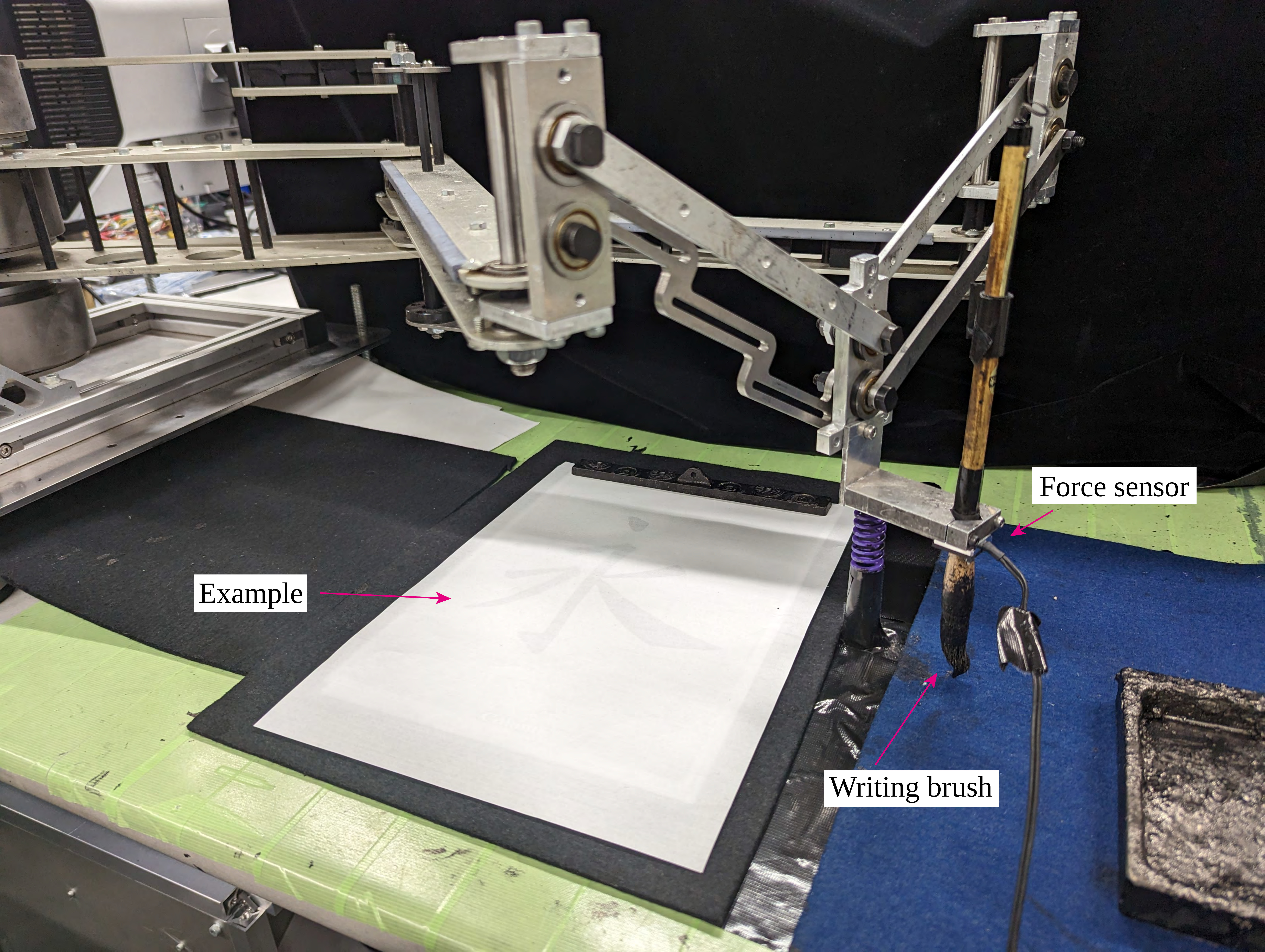}
    \caption{3-DOF calligraphy robot.}
    \label{fig:fuderobo}
  \end{center}
\end{figure}

\begin{figure*}[t]
    \begin{center}
      \includegraphics[width=15cm]{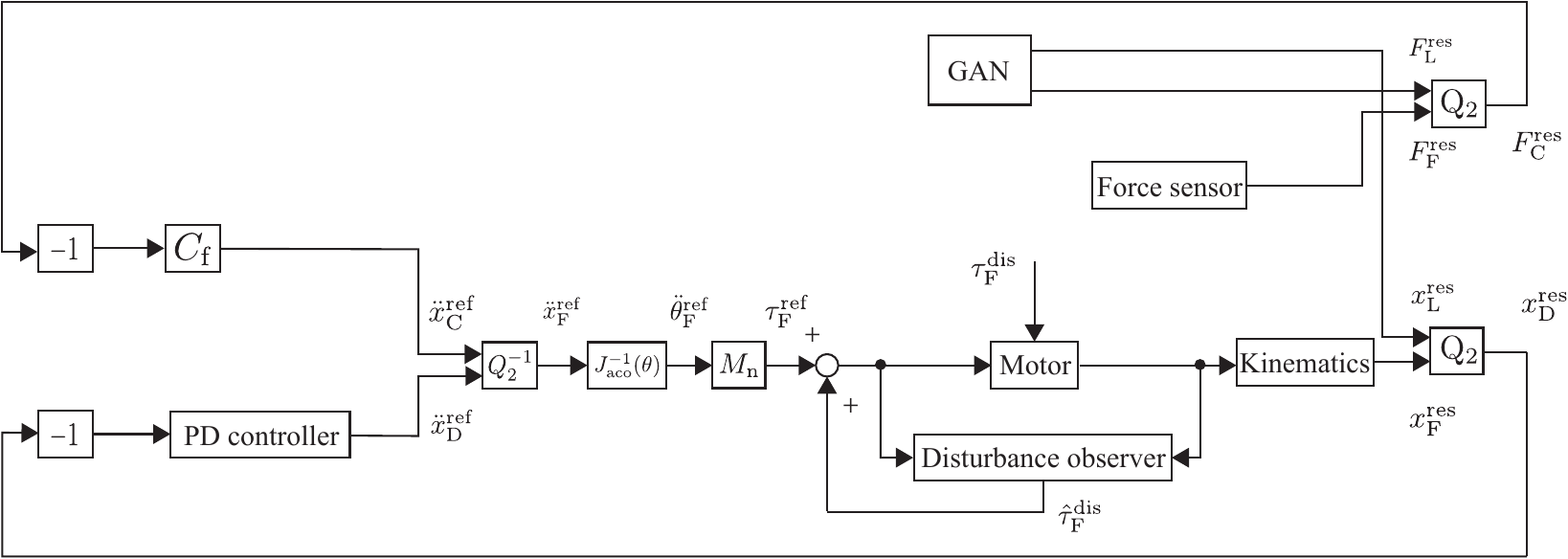}
      \caption{Block diagram of motion-copying system for calligraphy robot.}
      \label{fig:mcs}
    \end{center}
\end{figure*}

\begin{figure}[t]
    \begin{center}
      \includegraphics[width=5cm]{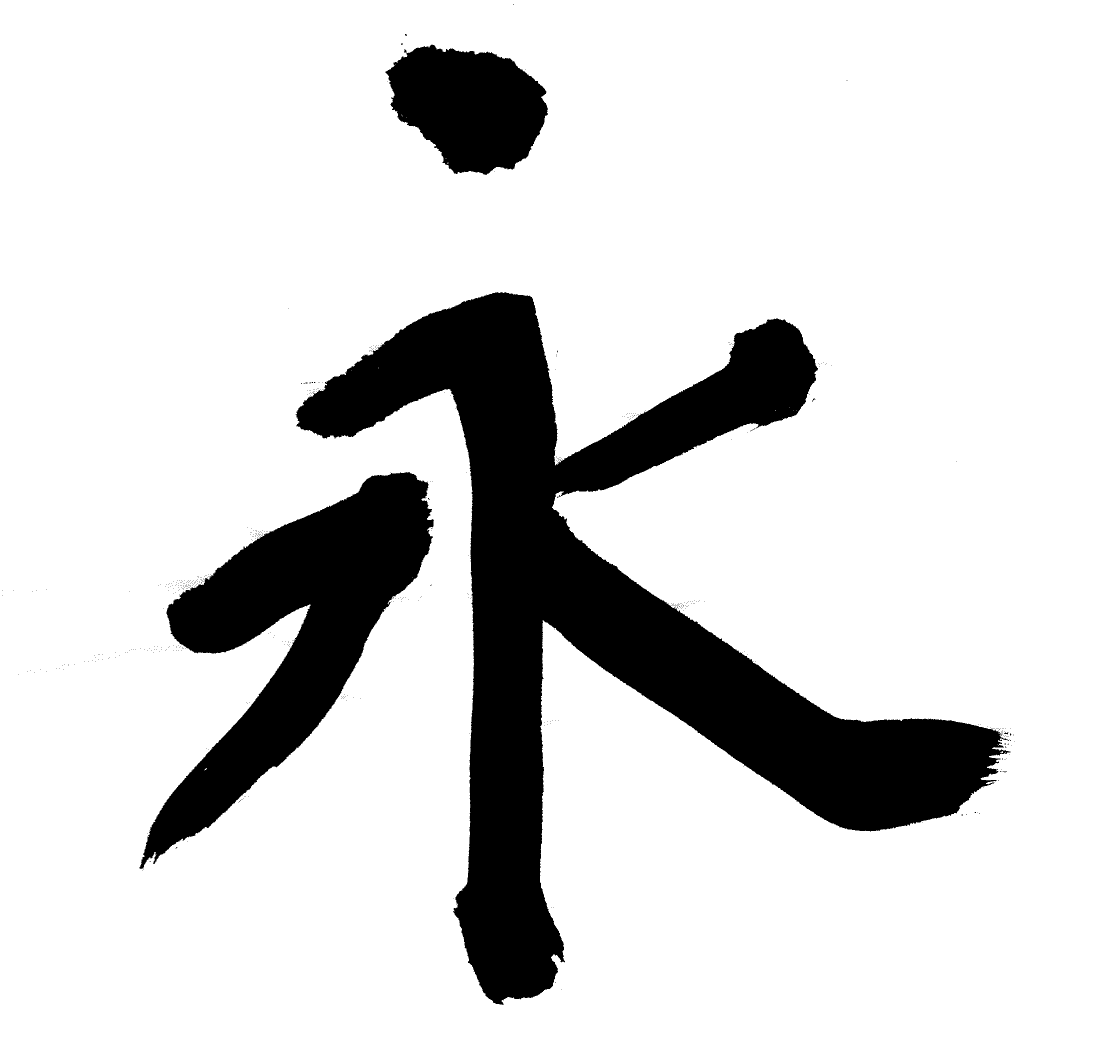}
      \caption{One of calligraphies by non-expert (Used as evaluation).}
      \label{fig:ne}
    \end{center}
\end{figure}

\begin{figure}[t]
    \begin{center}
      \includegraphics[width=5cm]{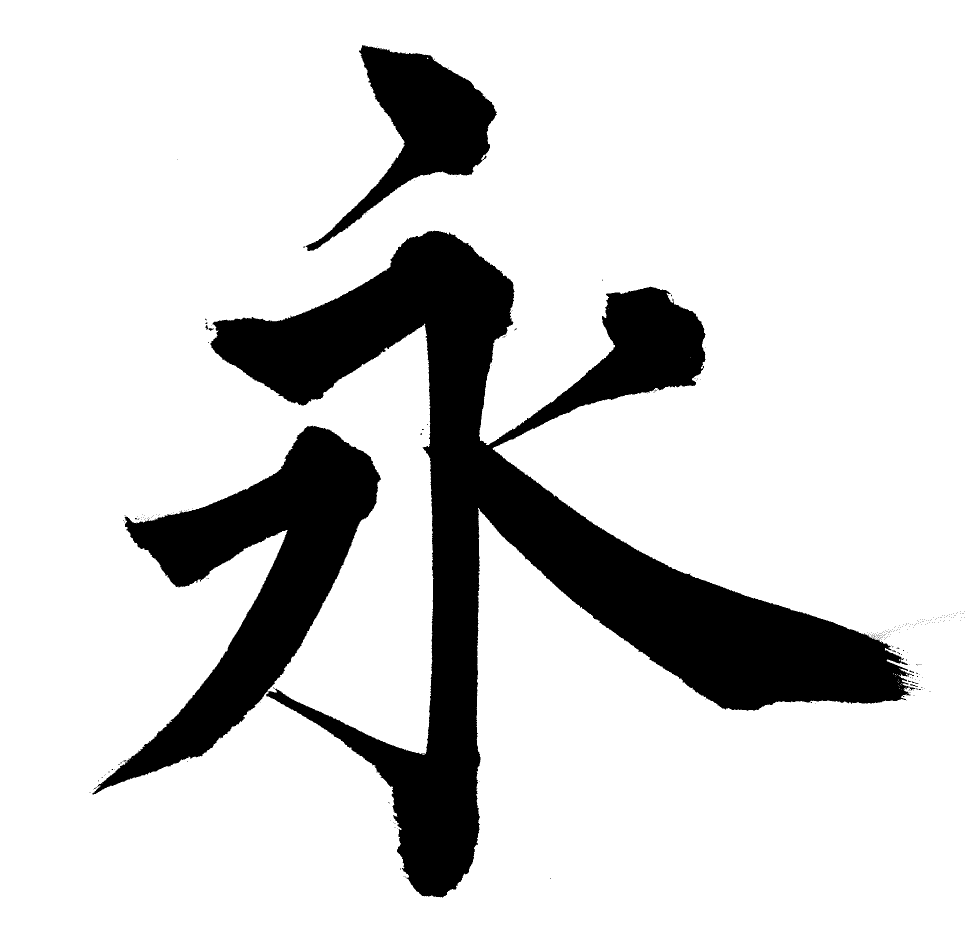}
      \caption{One of calligraphies by expert (Used as evaluation).}
      \label{fig:e}
    \end{center}
\end{figure}

\begin{figure}[t]
  \begin{center}
    \includegraphics[width=8.5cm]{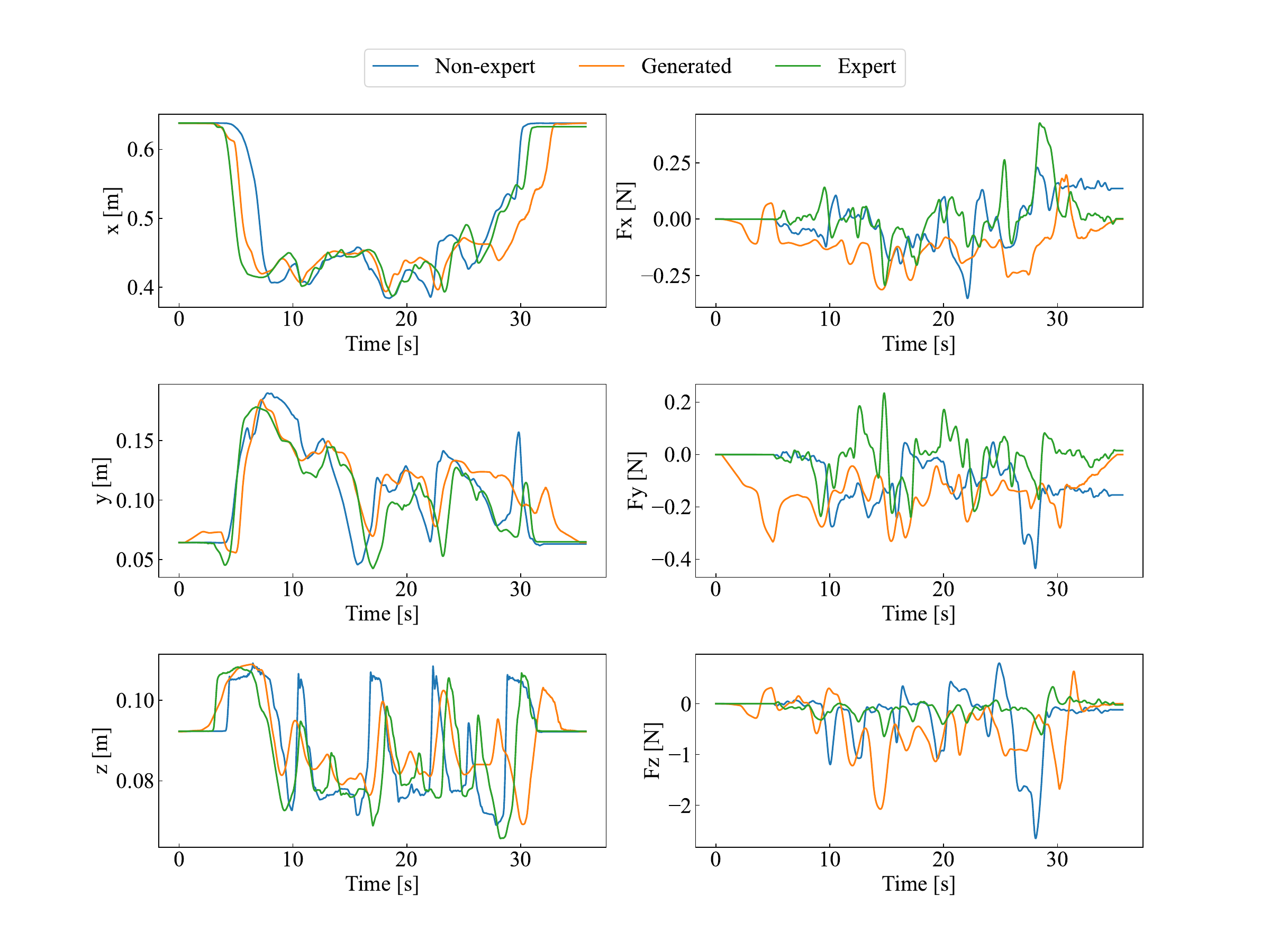}
    \caption{Comparison between expert/non-expert and generated motion data.}
    \label{fig:gen_motion}
  \end{center}
\end{figure}

\begin{figure}[t]
  \begin{center}
    \includegraphics[width=8cm]{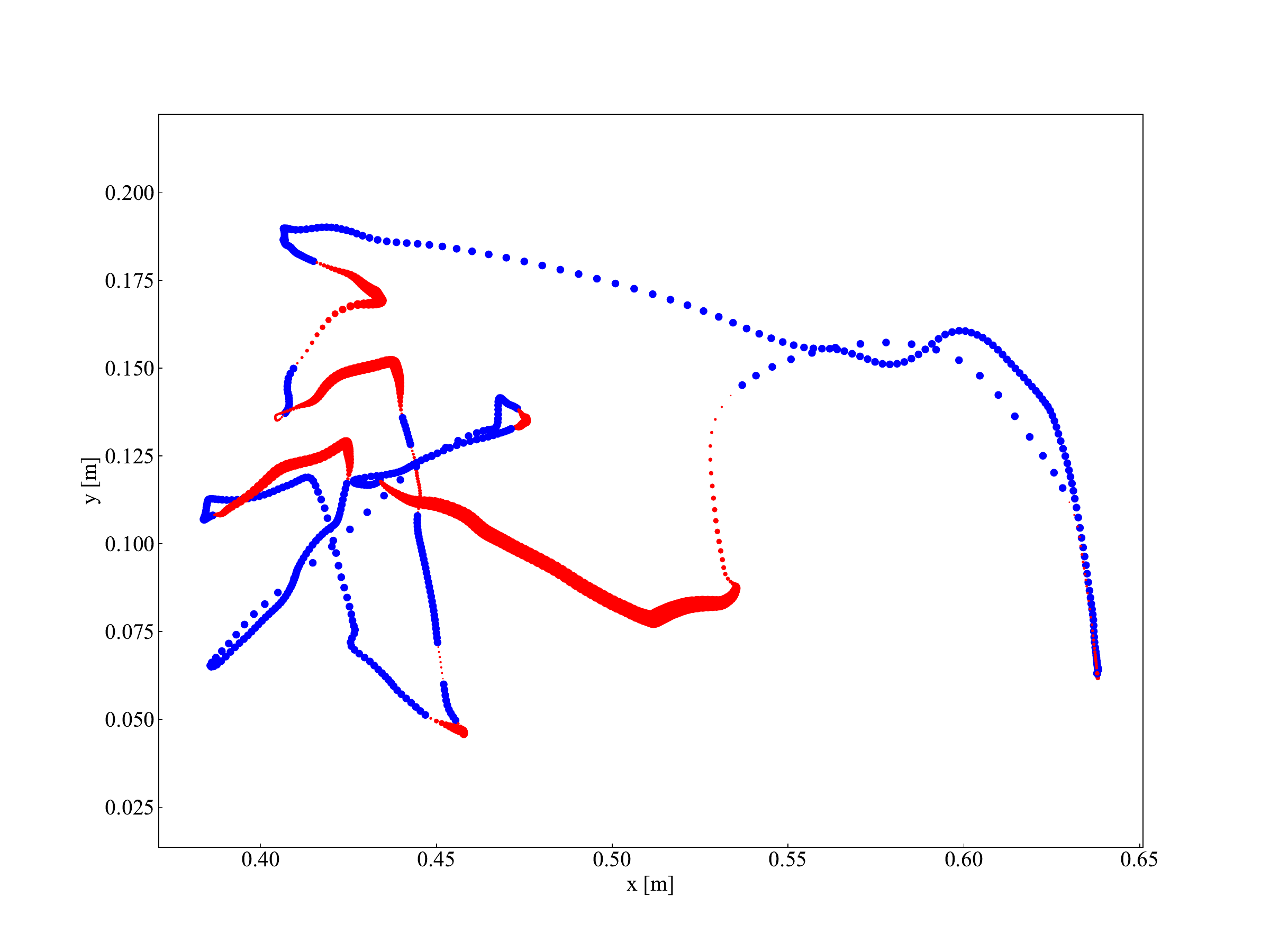}
    \caption{Trajectory of the non-expert data in the xy plane with force information.}
    \label{fig:xy_ne}
  \end{center}
\end{figure}

\begin{figure}[t]
  \begin{center}
    \includegraphics[width=8cm]{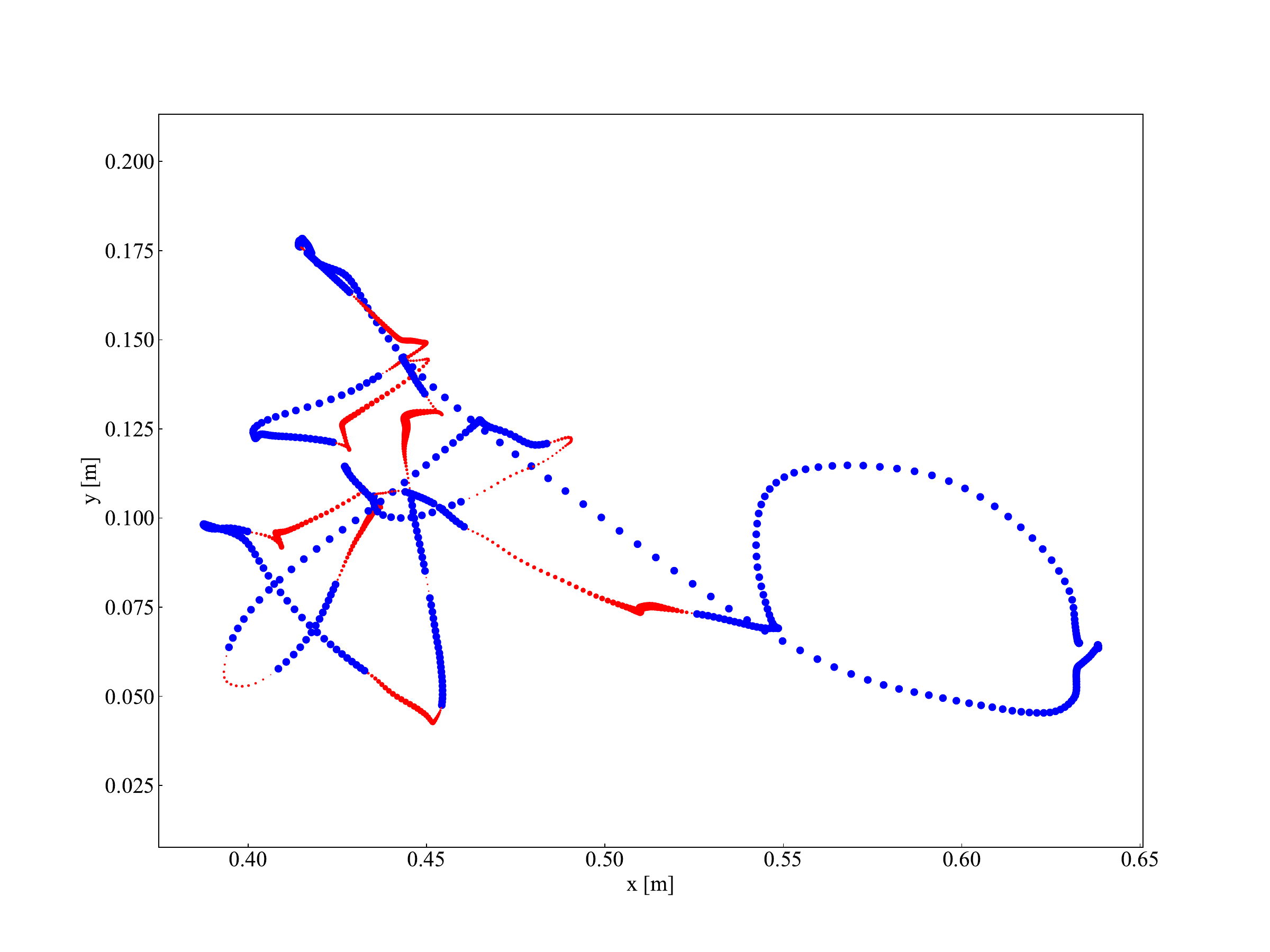}
    \caption{Trajectory of the expert data in the xy plane with force information.}
    \label{fig:xy_e}
  \end{center}
\end{figure}

\begin{figure}[t]
  \begin{center}
    \includegraphics[width=8cm]{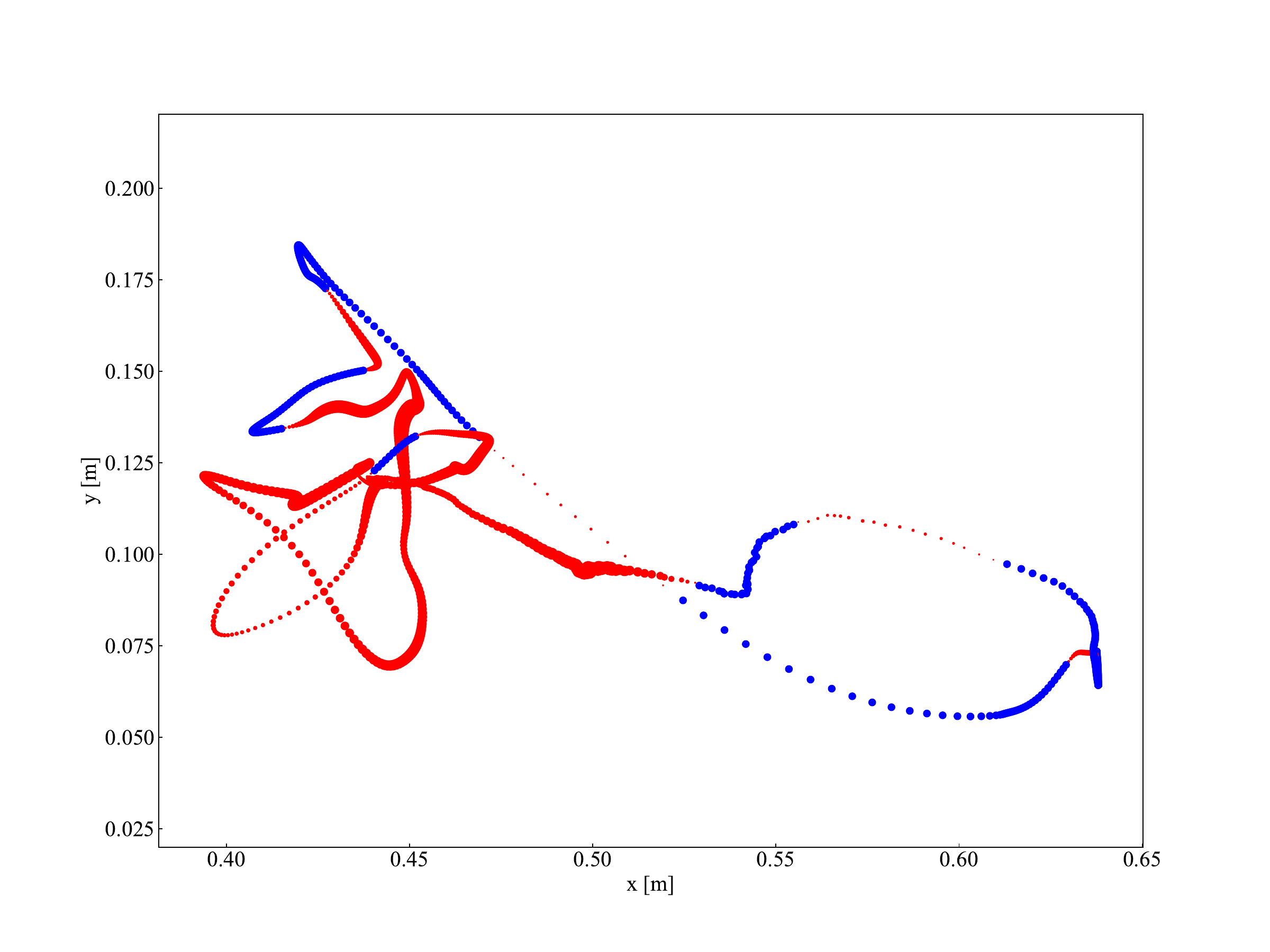}
    \caption{Trajectory of the generated data in the xy plane with force information.}
    \label{fig:xy_gene}
  \end{center}
\end{figure}

\begin{figure}[t]
  \begin{center}
    \includegraphics[width=6cm]{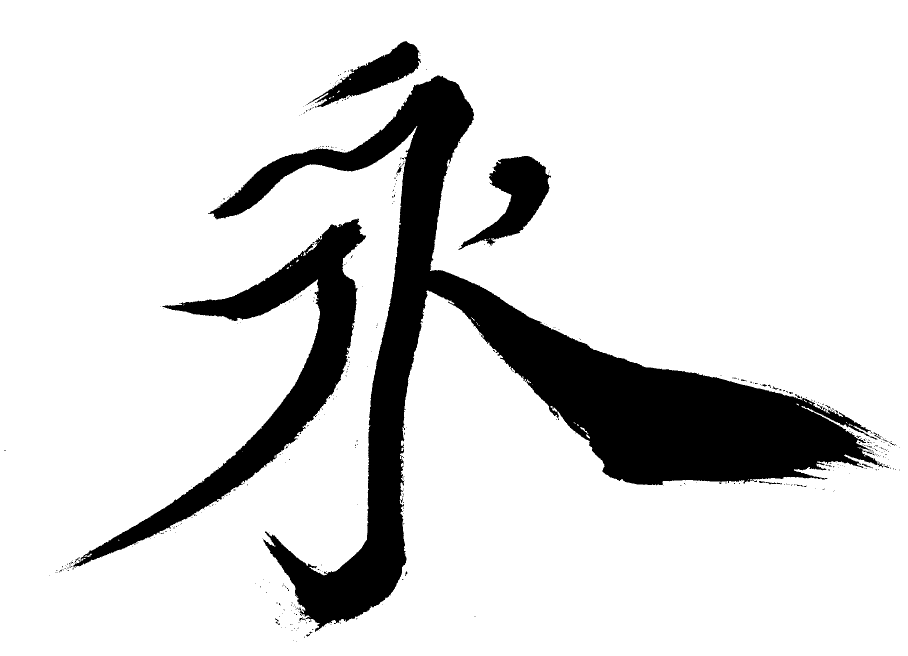}
    \caption{Calligraphy by robot using trained model.}
    \label{fig:robot_chara}
  \end{center}
\end{figure}

\begin{table}[t]
  \caption{Parameters of controller in the motion-copying system.}
  \label{table:mcs}
  \centering
   \vspace{5mm}
   \begin{tabular}{c|l|rc}
    \hline
    $T_s$ & Period of control system & 100 & $\mu$s\\
    $K_p$ & P gain of PD controller & 1200 & \\
    $K_d$ & D gain of PD controller & 70 &\\
    $g_{pd}$ & Pole of pseudo derivative & 40 & rad/s\\
    $g_f$ & Pole of low pass filter & 20 & rad/s\\
    \hline
   \end{tabular}
\end{table}

\begin{table}[t]
  \caption{Comparison of DTW distance between expert and non-expert/generated data.}
  \label{table:dtw}
  \centering
   \vspace{5mm}
   \begin{tabular}{c||rr}
    \hline
    Variables & Non-expert & Generated\\
    \hline \hline
    $x$ & 2.82 & 2.02\\
    $y$ & 1.60 & 1.35\\
    $z$ & 0.683 & 0.589\\
    $Fx$ & 7.07 & 9.95\\
    $Fy$ & 8.60 & 9.45\\
    $Fz$ & 31.7 & 35.0\\
    \hline
   \end{tabular}
\end{table}

\subsection{Experimental Setup}
To evaluate the proposed method, we generated brush-writing motion from expert's writing motion. Fig. \ref{fig:fuderobo} shows the 3-DOF calligraphy robot used in experiments. A expert and a non-experts used the brush with force sensor. The encoders saved position of brush through jacobian matrix. Motion-copying system in this robot is shown in Fig. \ref{fig:mcs}. The control parameters are shown in Fig.~\ref{table:mcs}. Generated motion data by GAN were used as saved leader data. A expert wrote three times, and a non-expert wrote six times. One of each data was used as evaluation data. To gather similar motion, cho-suiryo's work was used as an example character by putting a paper on the work. Expert's and non-expert's written images are shown in Figs. \ref{fig:ne} and \ref{fig:e}. To match a pair of motion data, we used DTW distance of $x$ and $y$ axis in this research because these values determine the rough shape of calligraphy. The sampling size of time series data was set to 800 (100 Hz x 8 s). $\lambda$ in Eq. \ref{loss_all} was set to 100, and we used ADAM as the optimization method of $G$ and $D$.

\subsection{Experimental Results}
Fig. \ref{fig:gen_motion} shows the comparison of non-expert/expert and generated motion data. Their waveforms are similar, however, generated force data had a relatively large error at both ends. Figs. \ref{fig:xy_ne}, \ref{fig:xy_e} and \ref{fig:xy_gene} are trajectories in the x-y plane with non-expert, expert and generated data, respectively. When $F_z$ is above 0.1 N, the line color was changed from blue to red. The trajectory at the bottom right is the round trip phase from the initial position of the writing brush. In generated motion data, $F_z$ was strengthened in the writing phase. We evaluated generated data by DTW distance against expert data. Results of DTW calculation between non-expert/generated and expert data are shown in Table \ref{table:dtw}. DTW distances in position are lower than non-expert data. In the force domain, however, DTW distances were higher. This result indicates that trained $D$ didn't use force information to identify the motion data true or fake. Fig. \ref{fig:robot_chara} shows written calligraphy by the 3-DOF calligraphy robot. The writing techniques of experts (``Hane'', ``Harai'') were observed in the generated calligraphy.

\section{Conclusions}

In this research, we proposed a novel motion translation method based on image translation using GAN. We succeeded in generating new expert-like motion data from non-expert data. The results indicate that this method enables users to teach robots tasks by inputting data, and skills from a trained model. In future work, we will use other networks to improve the errors in force domain. The goal of this research is to generate expert-like motion, thus, we need to carry out some sensory evaluations with generated calligraphies. 

\section*{Acknowledgments}
This work was partially supported by Strategic Information and Communications
R\&D Promotion Programme (SCOPE) Grant Number 201603011.

\bibliographystyle{IEEEtran}
\bibliography{ref}

\end{document}